\pdfoutput=1
\documentclass[11pt]{article}

\usepackage[margin=1.1in]{geometry}
\usepackage[T1]{fontenc}
\usepackage[utf8]{inputenc}
\usepackage{lmodern}
\usepackage{microtype}
\usepackage{booktabs}
\usepackage{longtable}
\usepackage{amsmath}
\usepackage[numbers,sort&compress]{natbib}
\usepackage{xcolor}
\usepackage[colorlinks=true,linkcolor=blue!60!black,citecolor=blue!60!black,urlcolor=blue!60!black]{hyperref}
\usepackage{caption}
\title{\textbf{Forty Questions: Identifying a Language Model\\from One-Word Answers}}
\author{Tapan Parikh\\Cornell Tech\\\texttt{tsp53@cornell.edu}}
\date{October 2026}

\begin{document}
\maketitle

\begin{abstract}
\noindent
A model behind an API can be identified from its one-word answers to everyday questions such as
``Name a sound'' or ``Name a poet''. A model's profile records how often it gives each answer to each
question. We attribute new answers to the model most likely to have given
them. On a panel of 94
models, 40 answers identify the exact model 92.5\% of the time, against 1.1\% by chance. Profiles
built through OpenRouter identified all ten models served by DeepInfra under blinded labels. A 
system prompt lowers accuracy by 6 to 16 points. We applied the method to 96 endpoints of 13 open
models on OpenRouter, and 92 identify as the model they claim. Of the other four, one matches the
claimed model with reasoning turned off. One serves Gemma~4 under the name Llama~4 Maverick: it
names itself Gemma~4, and it is billed Gemma~4's token counts on every message. Two remain
unresolved. Billing records also show
context added to first-party models. Grok~4.7 is billed 1,247 input tokens for a one-line
question that costs GPT-4o only 12 tokens, and five models state the current date without being billed for
it. The questions come from a public census of model defaults, and the method costs cents per
endpoint.
\end{abstract}

\section{Introduction}
\label{sec:intro}

A deployed language model is usually a name behind an API. The caller sees neither the weights
nor how they are served. A provider may quantize the weights, add a default system prompt or
change the reasoning setting, and each of these can change the answers without changing the
weights. In this paper we describe a method for the caller to tell which model is behind the API.

Recent work shows that short, constrained answers can identify a model. IRIS attributes endpoints
from requests for random digits and bit strings, with a trained classifier \citep{zhang2026iris}.
\citet{bruckner2026onetoken} verifies model identity from single-token answers to tasks such as
``name a random number''. FLIPS identifies configured instances from requested random binary
sequences \citep{richardeau2026flips}.

A census of model defaults asked 105 models 96 one-word questions such as ``Name a color'' and
``Name a fruit'', repeated eight times per question \citep{parikh2026census}. Models converge
strongly on these questions, as they do on open-ended prompts \citep{jiang2025hivemind}. The
answers that depart from the consensus repeat within a model. A model that says ``petrichor''
where the field says ``lavender'' says it again, and these repeated departures are what we use to
identify it. 

We use the 94 census models with at least eight answers on at least 90 of the 96
questions (\S\ref{sec:method}). We decide by likelihood
under per-question answer counts. No weights are tuned, and a new model is added by sampling it
once. Each decision traces to the answers behind it, so a human auditor can check it: which
answers the endpoint gave, and which models give them.

The method has
three common uses. An auditor checks whether an endpoint serves the model it claims
(\S\ref{sec:audit}). A buyer checks whether a vendor's model derives from another lab's
(\S\ref{sec:lineage}). A user checks whether the model behind a name has changed
(\S\ref{sec:weeks}). The auditor chooses the questions and sends no system prompt. A provider can still add text before
the auditor's message, and \S\ref{sec:audit} shows that this happens.

Contributions:
\begin{enumerate}\setlength{\itemsep}{0pt}
\item Closed-set identification of the exact model from one-word answers, with nothing learned
beyond answer counts. Forty answers identify one of 94 models 92.5\% of the time
(\S\ref{sec:howmany}).
\item Question selection. Ten questions selected on separate samples reach 81\% with two answers
each, against 67.5\% for ten random ones (\S\ref{sec:which}).
\item Blinded identification on a second host, and robustness to a system prompt and to repeated
sampling over weeks (\S\ref{sec:host} to \S\ref{sec:weeks}).
\item An audit of OpenRouter. Of 96 endpoints, 92 identify as the model they claim. Of the four
exceptions, one is consistent with a reasoning configuration, one is a substituted model and two remain
unresolved. A rescore against pinned reference profiles leaves these findings unchanged. Billed
tokens and the current date give evidence of context added to first-party models
(\S\ref{sec:audit}).
\end{enumerate}

\section{Related work}
\label{sec:related}

\paragraph{Random-sequence probes.} IRIS \citep{zhang2026iris} is the closest prior work. It sends
short random-generation requests (a coin draw, a digit, an integer in a range, a bit or digit
sequence) and classifies the responses with a random forest over 179 string features. On 17
OpenRouter endpoints it reaches 0.965 attribution accuracy at 8 queries. On 45 models it reaches
0.90 at 16 queries. It also flags 14 of 15 same-model OpenRouter provider pairs as
distinguishable. IRIS reports that a generic text classifier cannot attribute providers from
natural-language completions. Our answers are natural language, but each is one word to a fixed
question, so they can be counted.

\paragraph{Single-token fingerprints.} \citet{bruckner2026onetoken} samples 165 OpenRouter models
on ten one-token tasks in four languages. Jensen--Shannon distance between answer distributions
gives verification at a 7.3\% equal error rate with the full 40-cell battery, and family
assignment at 59.5\%. Bruckner reports no closed-set identification of the exact model. FLIPS
\citep{richardeau2026flips} uses biases in requested binary sequences. It treats a model under a
different system prompt or quantization as a different instance. It reports 96\% closed-set and
90\% open-set identification over 237 instances.

\paragraph{Query-based identification.} LLMmap \citep{pasquini2025llmmap} identifies 42 model
versions at over 95\% from eight crafted queries with a trained model. \citet{chen2026sysprompt}
show that a one-sentence tone prefix in the system prompt drops clone detection on short
structured outputs from 0.978 to 0.547.

\paragraph{Auditing APIs.} Model Equality Testing \citep{gao2025met} tests whether an API serves a
reference distribution with an MMD two-sample test, and finds 11 of 31 Llama endpoints serving
different distributions from Meta's weights. The rank-based uniformity test \citep{zhu2025rut}
detects quantization, fine-tuning and swaps from outputs. \citet{cai2025payfor} argue that
software-only checks are unreliable and propose trusted execution. Stability Monitor
\citep{leshin2026stability} detects changes in version, stack and quantization from output
distributions and finds stability differences across hosts of one model. AgentProv
\citep{wang2026agentprov} is the source of our check in \S\ref{sec:added}. Its token-count scan
finds 44\% of 75 OpenRouter open-weight endpoints off the official chat template, some with
injected system prompts. Under an injected prompt, MET and RUT falsely reject honest models 67\%
and 53\% of the time.

\paragraph{Serving nondeterminism.} Batch size varies with server load, and kernels that are not
batch-invariant then change outputs. At temperature zero, Qwen3 235B gave 80 distinct completions
in 1,000 \citep{he2025nondeterminism}. \citet{zhu2026unstable} find that byte-identical next-day
replays on shared endpoints agree at 0.78 in a preregistered LLM-judge study.

\section{Data and method}
\label{sec:method}

\paragraph{Questions.} 96 questions, 95 of them in the form ``Name a color. Reply with one word
only.'', such as ``Name a smell'' and ``Name a villain''. The other is ``Pick a word. Reply with
one word only.'' The full list is in Appendix~\ref{app:questions}.

\paragraph{Models.} 94 models with at least eight answers on at least 90 of the 96 questions,
carrying 27 family labels (Appendix~\ref{app:models}). All were queried through OpenRouter's
default routing, with no provider pinned, no system prompt, temperature 1 and default reasoning
settings, between 3 July and 2 October 2026. The robustness runs in \S\ref{sec:sysprompt} and \S\ref{sec:weeks} were made on 1 October, the provider audit in \S\ref{sec:audit} on 2 October, and its follow-ups on 6 and 11 October. Answers are normalized as in the census: first answer line, case-folded, spelling variants and plurals merged (Appendix~\ref{app:normalize}).

\paragraph{Profile and score.} A model's profile for question $c$ is the count of each answer among
its samples. A test set $\{(c, a)\}$ is scored under model $m$ as
\[
\sum_{(c,a)} \log \frac{n_m(c, a) + 0.1}{N_m(c) + 0.1\,V_c},
\]
where $n_m(c,a)$ is the count of answer $a$, $N_m(c)$ the number of profiled samples and $V_c$ the
number of distinct answers to $c$ across all models plus one. The prediction is the
highest-scoring model. This is a naive Bayes classifier with a pseudocount of 0.1. Nothing is fit
beyond the counts. When profiles have equal size, $V_c$ is shared by every candidate and does not
affect the ranking. Adding a model means sampling it once. Edge cases are in
Appendix~\ref{app:edges}.

\paragraph{Evaluation.} Four-fold over samples. Samples \{0,1\}, \{2,3\}, \{4,5\} and \{6,7\} are
held out in turn, and profiles use the other six, so no answer is scored against itself. Random
question subsets are drawn ten times per size.

\section{Identification}
\label{sec:ident}

\subsection{How many answers}
\label{sec:howmany}

\begin{table}[h]
\centering\small
\begin{tabular}{rcccc}
\toprule
& \multicolumn{2}{c}{one answer each} & \multicolumn{2}{c}{two answers each} \\
\cmidrule(lr){2-3}\cmidrule(lr){4-5}
questions & exact & family & exact & family \\
\midrule

5 & 28.1\% & 39.9\% & 41.5\% & 53.0\% \\
10 & 54.5\% & 63.6\% & 68.0\% & 76.2\% \\
20 & 78.2\% & 84.9\% & 90.7\% & 94.1\% \\
40 & 92.5\% & 96.0\% & 96.9\% & 99.1\% \\
96 & 98.4\% & 100.0\% & 98.7\% & 100.0\% \\

\bottomrule
\end{tabular}
\caption{Identification among 94 models by number of questions and answers per question.
Chance is 1.1\%.}
\label{tab:curve}
\end{table}

Accuracy depends mostly on the total number of answers (Table~\ref{tab:curve}). New questions
carry slightly more than repeats: twenty questions asked twice reach 90.7\%, and forty asked once
reach 92.5\%.

These held-out answers come from the same collection runs as the profiles they are scored
against. Answers collected on other days are identified at similar rates: 98\% for 14 models on a
later day with 31 questions asked twice (\S\ref{sec:sysprompt}) and 10 of 10 on a second host
(\S\ref{sec:host}).

\paragraph{Against simpler scorers.} We compared the likelihood with two simpler scorers on the
same folds and question draws. One picks the model whose answers are closest in Jensen--Shannon
distance, as in \citet{bruckner2026onetoken}. The other counts how many held-out answers match
each model's favourite answer. On the 66 questions every model answered eight times, forty
answers identify the model 92\% of the time under the likelihood, 65\% under Jensen--Shannon
distance and 76\% under favourite answers. The likelihood scores best at every budget
(Appendix~\ref{app:scorers}). With one or two held-out answers per question, the distance between
distributions cannot be estimated well. Matching favourite answers ignores the occasional answers
that distinguish models most.

\subsection{Which questions}
\label{sec:which}

Greedy selection adds, one at a time, the question that most raises the probability given to the
true model. To keep selection apart from scoring, questions are chosen on half of each model's
samples and scored on the other half. The halves are then swapped, and Table~\ref{tab:selection}
averages the two. Selected and random sets both come from the 66 questions every model answered
eight times.

\begin{table}[h]
\centering\small
\begin{tabular}{rcc}
\toprule
questions (two answers each) & selected & random \\
\midrule

5 & 62.0\% & 38.8\% \\
10 & 81.1\% & 67.5\% \\
15 & 90.7\% & 80.6\% \\
20 & 95.8\% & 87.9\% \\
25 & 97.1\% & 92.7\% \\

\bottomrule
\end{tabular}
\caption{Selected against random question sets, mean of the two halves. Both halves' choices are
in Appendix~\ref{app:selection}.}
\label{tab:selection}
\end{table}

Selection beats random sets at every size, by 14 points at ten questions. The questions it picks
are less stable. The two halves share four of their first ten: bird, dessert, landmark and
reptile. Many questions carry similar information, so which ones win depends on the samples.
Selection is tuned to this panel, so its advantage on unseen models is likely smaller.

\subsection{A second host, blinded}
\label{sec:host}

Ten panel models were queried through DeepInfra under random code names, two samples per question:
Llama 3.3 70B, Llama 4 Scout, Mistral Small 3.2, Mistral Nemo, Nemotron 3 Super, Qwen3 235B,
GPT-OSS 120B, DeepSeek V4 Flash, DeepSeek V3.2 and Gemma 3 27B. The scorer recorded its guesses
before the key was read. Profiles came only from OpenRouter answers. DeepInfra is one of
OpenRouter's providers, and the profiles' providers were not recorded (\S\ref{sec:method}), so some
reference answers may have come from DeepInfra. With all questions (93 to 96 per model), 10 of 10 were
identified (95\% interval 69\% to 100\%). With the best twenty, 8 of 10 were (44\% to 97\%). In
one miss the true model ranked second, 0.3 nats behind in total over 40 answers, which is a near
tie. In the other it ranked third, 5.9 nats behind.

Equality tests can still tell hosts of one model apart \citep{bruckner2026onetoken,zhang2026iris}.
Two hosts can differ slightly and still both be closest to the same model.

\subsection{Under a system prompt}
\label{sec:sysprompt}

Fourteen models answered 31 of the 96 questions three ways: with no system prompt, with ``You are a
helpful assistant.'', and with a made-up persona, ``You are Ava, a friendly assistant.''. They were
identified among all 94 models from profiles built with no system prompt.

\begin{table}[h]
\centering\small
\begin{tabular}{lcccc}
\toprule
condition & 10 $\times$ 2 & 20 $\times$ 2 & 31 $\times$ 2 & 31 $\times$ 4 \\
\midrule

no system prompt, a later day & 66\% & 90\% & 98\% & 100\% \\
``You are a helpful assistant.'' & 60\% & 82\% & 89\% & 94\% \\
``You are Ava, a friendly assistant.'' & 56\% & 74\% & 86\% & 91\% \\

\bottomrule
\end{tabular}
\caption{Identification of 14 models under a system prompt, among all 94 profiles (20 draws per
cell).}
\label{tab:sysprompt}
\end{table}

The generic prompt costs 6 to 9 points, and the Ava persona costs 9 to 16. With 31 questions asked four times, models under a system prompt are still identified
91\% to 94\% of the time. A provider can add such a prompt without the auditor knowing
(\S\ref{sec:added}).

\subsection{Over weeks}
\label{sec:weeks}

The fourteen models' favourite answers in the robustness run matched their original runs, made
weeks to three months earlier, on 89\% of questions. Two halves of the same day's samples matched
in 84\%, with fewer samples per half. On this coarse measure the fingerprint is stable over the
collection period.

\subsection{A lineage}
\label{sec:lineage}

Writer markets Palmyra X5 as its own model and has not named a base model for it. Writer's report
on the successor, Palmyra x6, names GLM-5.2 as that model's base \citep{du2026palmyrax6}. Palmyra
X5's nearest model on our 96 questions is Qwen3 235B Instruct (2507), by 2.3 nats per question over
the next candidate. Qwen3 235B's nearest model is Palmyra X5. \citet{bruckner2026onetoken} found the
same pair at a distance equal to the median distance between halves of one model. Mutual nearest
neighbours show that the two answer alike. They do not prove that one is derived from the other.
Palmyra X5 is also billed 243 input tokens for a one-line question (\S\ref{sec:added}), so added
context may shape its answers.

\section{Validation: an audit of OpenRouter providers}
\label{sec:audit}

\subsection{Setup}

OpenRouter routes each open model to one of several providers, with declared precision (bf16,
fp8, fp4, int4 or unstated) and prices that differ up to thirtyfold for one model. We pinned each
of 104 endpoints of 14 open models in turn, with fallbacks off, and asked all 96 questions four
times. Eight could not be scored. Four returned no data, and four claim Qwen3.5 122B, which is not in the panel because it answered fewer than 90 questions eight times. The rest of this
section reports the other 96 endpoints, of 13 models. The whole audit cost \$7.20 over 36,918
calls. The released transcripts identify each endpoint by provider. The text characterizes
endpoints and names no third-party provider; Appendix~\ref{app:audit} lists them by model and
letter.

Each endpoint is scored two ways. Identification ranks the 94 panel profiles by the likelihood of
the endpoint's answers, and reports where the claimed model ranks. A peer test compares the
endpoint with the pooled answers of the same model's other providers, collected the same day. The
distance is the mean Jensen--Shannon divergence across questions. Its $p$-value comes from 1,000
label permutations within each question, with Benjamini--Hochberg correction across endpoints.

\subsection{Results}
\label{sec:results}

Of the 96 endpoints, 92 identify as the model they claim. Table~\ref{tab:mismatch} lists the four
that do not.

The peer test flags 22 endpoints as different from the same model's other providers at
$q < 0.05$. These include all four mismatches. The other 18 still identify as the claimed model,
so they differ from their peers within the claimed model. The reasoning setting is one source of
difference. DeepSeek V4 Flash, for example, reasoned on at least 99\% of calls at 11 of its 15 endpoints, on no call at
three, and on 38\% of calls at one. The peer pools include the mismatched endpoints, so their peers
can be flagged too. Both Maverick endpoints that identify as Maverick have $q = 0.006$.

\begin{table}[h]
\centering\small
\begin{tabular}{llllr}
\toprule
claimed model & declared precision & identified as & rank of claim & $q$ \\
\midrule
Llama 4 Maverick & fp8 & Gemma 4 31B & 91 & 0.006 \\
Mistral Small 3.2 & unstated & Mistral Small 2603 & 5 & 0.006 \\
Kimi K2.5 & unstated & Kimi K2 & 2 & 0.006 \\
DeepSeek V4 Flash & fp4 & Qwen 3.8 27B & 4 & 0.048 \\
\bottomrule
\end{tabular}
\caption{The four endpoints that do not identify as claimed. Rank is the claimed model's rank
among 94 profiles.}
\label{tab:mismatch}
\end{table}

The Llama 4 Maverick endpoint is identified as Gemma 4 31B and ranks Maverick 91st of 94, so its
answers are far from anything Maverick says. Section~\ref{sec:followup} confirms that it serves
Gemma 4: it names itself Gemma 4, and it is billed Gemma 4's token counts.

The other three rank their claimed model near the top. The Mistral Small 3.2 endpoint is identified as Mistral Small 2603, a later release
from the same developer, and ranks the claimed model fifth. The Kimi K2.5 endpoint is identified as
Kimi K2, the earlier version of the same model, and ranks the claimed model second. The DeepSeek V4
Flash endpoint is identified as Qwen 3.8 27B, a model from another lab, and ranks the claimed model
fourth. The Mistral and DeepSeek identifications change between runs (\S\ref{sec:followup}), so
neither should be read as the model served.

\paragraph{How far a match is from a mismatch.} The margin is the claimed model's score minus the
best other model's, in nats per question. It is positive when the endpoint is identified as
claimed. Over the 92 matching endpoints the median margin is $+1.6$, the lowest three are $+0.12$,
$+0.20$ and $+0.24$, and 95\% are above $+0.47$. The four mismatches sit at $-7.0$ (Llama 4
Maverick), $-2.4$ (Mistral Small 3.2), $-0.9$ (Kimi K2.5) and $-0.25$ (DeepSeek V4 Flash). The
DeepSeek V4 Flash margin is about as far below zero as the weakest matches are above it, so on the
fingerprint alone it is a near miss.

\paragraph{Mixed reference profiles.} The panel profiles were collected through default routing
(\S\ref{sec:method}), so each may include answers from the endpoints this audit flags. We
estimated the share for each mismatch with a two-part mixture fitted to the panel profile's
answers. One part is the mismatch's answer distribution, and the other is the pooled answers of
the same model's matching endpoints. The share is 0.5\% for the Gemma 4 endpoint in the Llama 4
Maverick profile, 2\% for Kimi K2.5 and 10\% for DeepSeek V4 Flash. For Mistral Small 3.2 it is
17\%, most of it in the second four samples of the first 31 questions, collected on 30 September
(64\% of those answers). The mismatched Mistral endpoint is the developer's own API, which default
routing would often reach.

We then rescored every endpoint with its claimed model's profile replaced by a pinned one: eight
answers per question drawn from the same model's other matching endpoints on the audit day, the
endpoint itself left out. The four mismatches stay mismatches, with the claimed model ranked 92nd,
10th, 2nd and 7th. Of the 91 matching endpoints that have a matching peer, 90 still identify as
claimed. The exception is a GLM 4.7 endpoint that falls to second, behind Qwen 3.7 Plus, by 0.09
nats per question. Mixing in the reference profiles therefore does not explain the mismatches,
and it does not create the matches.

\subsection{Added prompts}
\label{sec:added}

An endpoint can serve the claimed weights and still answer differently, because text was added
before the auditor's message. AgentProv shows that this makes equality tests reject honest models.
Section~\ref{sec:sysprompt} shows it lowers identification accuracy. A provider pays to process
added text on every call, so billed input tokens above the count for the sent message reveal it.

\paragraph{Hidden context in first-party models.} All calls in this section went through
OpenRouter, so text added by OpenRouter cannot be told apart from text added by the lab. On the
audit day we asked every live panel model, with no system prompt, ``What is today's date?''. A
model told nothing can only guess the date. GPT-5, GPT-5.4, GPT-5.4 mini and GPT-5.5 stated the
correct date in both samples. Each was billed 11 input tokens for the question. GPT-5.6 and the
GPT-6 models were billed the same 11 tokens and did not know the date. Muse Spark 1.3, from
another lab, also stated the date in both samples while billed 12 tokens. The date reaches these five models without being billed, either as unbilled prompt text or through another channel such as server-side metadata. Either way, billed tokens are a
lower bound on added context.

Other models are billed for large added prefixes (Table~\ref{tab:billed}). Tokenizers differ, so
the GPT-4o count is a reference and not an exact baseline.

\begin{table}[h]
\centering\small
\begin{tabular}{lr}
\toprule
model & billed input tokens \\
\midrule
GPT-4o (reference) & 12 \\
Grok 4.7 & 1,247 \\
Palmyra X5 & 243 \\
Grok 4.5 and 4.6 & 211 \\
MiniMax M3 & 181 \\
Kimi K3 & 90 to 93 \\
\bottomrule
\end{tabular}
\caption{Billed input tokens for ``What is today's date?'', no system prompt.}
\label{tab:billed}
\end{table}

Grok 4.7's surplus over GPT-4o is 1,235 tokens on every message we sent, from a census question
to a request to quote its instructions. Reasoning is billed separately as
output and varies from 29 to 498 tokens per call, while the input count stays fixed. From the
second call on, 1,152 input tokens per call came from cache, so the prefix is the same text every
time. We cannot see its content. 

\paragraph{The billed-token check on the audit.} For each endpoint we compared billed input tokens
with the most common count among the same model's other providers for the same message. Peers
share a tokenizer, so a difference measures added text or a different chat template. Of the 95
scored endpoints that have peers, 19 are billed above their peers. Fifteen of the 19 identify as
the claimed model, including three Llama 3.3 70B endpoints at $+25$ tokens and two DeepSeek V4
Flash endpoints at $+79$. All four mismatches are among the 19.

Four days after the audit we sent three messages to each mismatch and to one or two of its peers:
the census question, the same with our own one-line system prompt, and a longer user message
(Table~\ref{tab:surplus}).

\begin{table}[h]
\centering\small
\begin{tabular}{lcccc}
\toprule
& & \multicolumn{3}{c}{four days later} \\
\cmidrule(lr){3-5}
claimed model & audit & plain & our system prompt & longer message \\
\midrule
Llama 4 Maverick & $+3$ & $+3$ & $+4$ & $+3$ \\
Mistral Small 3.2 & $+12$ & $+12$ & $+12$ & $+12$ \\
Kimi K2.5 & $+19$ & $+19$ & $+1$ & $+19$ \\
DeepSeek V4 Flash & $+79$ & $+79$ & $+79$ & $+79$ \\
\bottomrule
\end{tabular}
\caption{Billed input tokens above the peers' count, per mismatch.}
\label{tab:surplus}
\end{table}

The Mistral Small 3.2, Kimi K2.5 and DeepSeek V4 Flash surpluses are constant across message
lengths, so each is a fixed prefix. The Kimi K2.5 prefix disappears when the caller sends a system
prompt, which is consistent with a default system prompt that the caller's replaces. The other two
persist. 

\subsection{Follow-up runs on the four mismatches}
\label{sec:followup}

Four days after the audit we ran three follow-ups, each with all 96 questions four times per
endpoint.
\begin{enumerate}\setlength{\itemsep}{0pt}
\item \textbf{A later day.} The four mismatches again, with no system prompt.
\item \textbf{A caller's system prompt.} Every endpoint of the four models, mismatches and peers,
under ``You are a helpful assistant.''. On many serving stacks a caller's system prompt replaces a
provider's default one. We assume it does here, and the Kimi K2.5 surplus falling to $+1$ shows it
on one endpoint.
\item \textbf{Reasoning off.} Every endpoint of Kimi K2.5 with reasoning turned off. The Kimi K2.5
mismatch answered with a median of 3 output tokens and no reasoning, where its peers reasoned on
every call.
\end{enumerate}

\begin{table}[h]
\centering\small
\begin{tabular}{lllll}
\toprule
claimed model & audit & later day & caller's system prompt & surplus \\
\midrule
Llama 4 Maverick & Gemma 4 31B (91) & Gemma 4 31B (91) & Gemma 4 31B (91) & $+4$ \\
Mistral Small 3.2 & Mistral Small 2603 (5) & Mixtral 8x22B (3) & Mistral Small 2603 (5) & $+12$ \\
Kimi K2.5 & Kimi K2 (2) & Kimi K2 (2) & Kimi K2 (2) & $+1$ \\
DeepSeek V4 Flash & Qwen 3.8 27B (4) & Grok 4.5 (7) & Grok 4.5 (9) & $+79$ \\
\bottomrule
\end{tabular}
\caption{Identification of the four mismatches across runs, with the claimed model's rank in
parentheses. Surplus is billed input tokens above the peers under the caller's system prompt.}
\label{tab:followup}
\end{table}

None of the four endpoints came out as its claimed model on any of the three runs. Under the
caller's system prompt, 21 of the 22 peers that returned answers identify as their claimed model.
The exception is a DeepSeek V4 Flash peer whose claimed model ranks second. All four mismatches
differ from their peers at $q < 0.05$ under the system prompt.

\paragraph{Kimi K2.5.} Under the caller's system prompt its billed surplus falls from $+19$ to
$+1$, so a default prompt was removed, and it still identifies as Kimi K2. With reasoning off, all
three peers that returned one-word answers also identify as Kimi K2, the non-reasoning
predecessor, with the claimed model second. The fourth peer returned its reasoning as the reply
when reasoning was off and could not be scored. The mismatch is therefore consistent with the
claimed model served with reasoning off. The fingerprint cannot separate this from Kimi K2 served
under the K2.5 name, since both give the same answers. We count it as a likely configuration
difference.

\paragraph{A disclosure probe on the other three.} Each of the other three keeps its billed surplus
under the caller's system prompt. A two-sample test such as Model Equality Testing would only
confirm what the peer test shows, that they differ from their peers. We asked instead what the
endpoints say about themselves. Each mismatch and two of its peers received three messages with no
system prompt, three samples each: ``What is today's date?'', a request to quote any instructions
given before the message, and ``What is your name, and which company made you?''. 

\paragraph{Llama 4 Maverick.} The endpoint named itself Gemma 4, developed by Google DeepMind, in
two of three samples, and said it was trained by Google in all three answers about its
instructions. Both peers named Llama and Meta in every sample. The endpoint was billed 20, 43 and
24 input tokens on the three messages. Gemma 4 31B through its own providers is billed exactly 20,
43 and 24 on the same messages, and Llama 4 Maverick 15, 40 and 21. The surplus is the difference
between two tokenizers. The fingerprint, the model's account of itself and the tokenizer agree, so
we count this endpoint as Gemma 4 served under the name Llama 4 Maverick.

\paragraph{DeepSeek V4 Flash.} The endpoint reasoned for 2,100 to 3,800 tokens on the date and
instruction questions, in numbered markdown steps. Two other providers are billed the same 79-token
surplus and identify as the claimed model (\S\ref{sec:added}). The one in this probe reasoned for 42 to 485 tokens, in the terse style
of every other DeepSeek V4 Flash endpoint. If the two prefixes are the same text, the prefix does
not cause the long reasoning, though equal counts do not prove equal text. While reasoning about
its instructions, the endpoint listed ``You are a large language model, trained by Google'' and
asked itself whether it may call itself Gemma. Its replies named DeepSeek, and its billed counts
fit the claimed model's tokenizer plus the surplus. Models confabulate their instructions, so the
reasoning text is weak evidence alone. The evidence points away from the claimed weights, but the served model remains unresolved.

\paragraph{Mistral Small 3.2.} This endpoint is the developer's own API. It reported no
instructions, identified itself as Mistral AI, and gave invented dates from 2024 and 2025. Its billed counts (21, 45
and 26) are the developer API's counts for both Mistral Small 3.2 and Mistral Small 2603, which
share a tokenizer and template. Nothing in the probe separates two releases from one developer.
The fingerprint places the endpoint nearest Mistral Small 2603 on two of three runs. That is
consistent with the developer serving a newer release under the older name. It is also consistent
with the claimed weights under the endpoint's fixed 12-token prefix, which persists under the
caller's system prompt. It also remains unresolved.

The 92 matches show that the fingerprint identifies models as a live service serves them. The Kimi
K2.5 and Llama 4 Maverick cases show two things a mismatch can mean: a configuration and another
lab's model.

\section{Limitations}
\label{sec:limits}

\begin{enumerate}\setlength{\itemsep}{2pt}
\item \textbf{Serving nondeterminism.} Outputs from one endpoint vary with load, batch size and
hardware \citep{he2025nondeterminism}. Byte-identical next-day replays agree at 0.78 in a
preregistered study of shared endpoints \citep{zhu2026unstable}. Nondeterminism moves token
probabilities most in the range 0.1 to 0.9 \citep{fu2026nondeterminism}, which is where contested
answers lie. Our stability measure in \S\ref{sec:weeks} is coarse: it checks only each model's
favourite answer.
\item \textbf{Closed set.} The method picks among profiled models. An unseen model is assigned to
its nearest relative. Open-set rejection is untested.
\item \textbf{Added context and routing.} Some first-party models and some providers receive text
before the caller's message (\S\ref{sec:added}). Our profiles were built through OpenRouter's
default routing, so they include whatever was added there. A profile describes a model as served.
It also mixes providers. On the audit day four of ten unpinned calls to Llama 4 Maverick were
billed the token counts of the endpoint serving Gemma 4, but the Maverick profile was collected in July and September and
holds about 0.5\% Gemma 4 answers. The Mistral Small 3.2 profile is about 17\% from the developer's
own endpoint (\S\ref{sec:results}). Pinned reference profiles leave the audit's conclusions
unchanged, but the identification results in \S\ref{sec:ident} rest on mixed profiles.
\item \textbf{Billing is a lower bound.} Five models received the date with no billed tokens
(\S\ref{sec:added}). Context that is not billed cannot be found by the billed-token check.
\item \textbf{Within-family confusion.} Most errors at small budgets fall outside the true family.
As answers grow, the remaining errors move inside it, and at 96 questions all of them are within
one family.
\item \textbf{Reasoning setting.} Profiles were built at each model's default reasoning setting. A
hybrid model served with reasoning off can resemble a different model, as Kimi K2.5 resembles
Kimi K2 (\S\ref{sec:followup}). Profiling hybrid models in both settings would close this gap.
\item \textbf{Temperature.} All data are at temperature 1.
\item \textbf{No matched baseline from prior work.} We did not run random-sequence probes on this
panel, so any comparison with IRIS and FLIPS rests on their published numbers.
\item \textbf{Gaming.} A provider who knows the public questions could route them to the genuine
model, add an answer key in a hidden prompt, or fine-tune a cheaper model to imitate the answers.
A rotating secret reserve of questions could mitigate the first. The billed-token check could
catch the second only when the added text is billed (limitation 4). None of this is tested. Many
questions carry similar information (\S\ref{sec:which}), so a reserve of new questions may work as
well as the published ones. Random sets of 40 questions already identify the model 89\% to 94\% of
the time, whichever 40 are drawn (Appendix~\ref{app:scorers}).
\end{enumerate}

\section{Conclusion}

One-word answers to everyday questions identify the model behind an API. Forty answers pick one of
94 models 92.5\% of the time with nothing learned beyond answer counts, and they identify all ten
models tested on a second host. Each decision can be checked by reading the answers behind it. Applied to 96
OpenRouter endpoints, the method identifies 92 as claimed. One of the four exceptions matches the
claimed model served with reasoning off. Another is Gemma 4 served under the name Llama 4 Maverick.
The other two remain unresolved. Billed tokens and the date show that first-party models served
through OpenRouter also receive context the caller did not send.

\section*{Disclosure}

We sent the Llama 4 Maverick finding to the provider and to OpenRouter on 14 October 2026, before
posting. The endpoint was still billed Gemma 4's token counts that day. The provider replied that
the endpoint serves Meta's official FP8 release of Llama 4 Maverick, and did not address the token
counts. OpenRouter matched our requests to that endpoint in its records and reproduced the token
counts on 15 October. On 20 October the endpoint was no longer listed on OpenRouter, and requests pinned to it
returned not found. The paper names no third-party provider, and the released transcripts do.

\section*{Data and code}

Everything is in \url{https://github.com/tap2k/modelun}, under \texttt{studies/consensus/}, at the
tag \texttt{consensus-fingerprint-arxiv-v1}.
\begin{itemize}\setlength{\itemsep}{0pt}\raggedright
\item Identification: \texttt{probe\_fingerprint.py}, \texttt{probe\_fingerprint\_baselines.py},
\texttt{probe\_perturb.py} and \texttt{probe\_fingerprint\_robust.py}, with outputs in
\texttt{probes/}.
\item Audit: \texttt{probe\_provider\_audit.py}, \texttt{probe\_billed\_tokens.py},
\texttt{run\_provider\_followup.py}, \texttt{probe\_provider\_followup.py},
\texttt{probe\_reference\_contamination.py} and \texttt{probe\_disclosure.py}. The transcripts in
\texttt{transcripts-providers/} identify each endpoint by provider.
\item Blind test: \texttt{probes/blind-deepinfra-2026-10-02.json}. Date and billing probe:
\texttt{hidden-prompt-2026-10-02/}.
\item Tables: \texttt{fingerprint\_paper.py}.
\end{itemize}

\section*{Note on AI usage}
This work was done with Claude (Opus 5.5), which helped run the probes, build the analysis and
draft the text; the research questions and interpretation are the author's. Several Claude
models are also subjects of the study.

\bibliographystyle{plainnat}
\bibliography{references}

@misc{bruckner2026onetoken,
  author = {Bruckner, T.},
  title = {One Token Is Enough: Fingerprinting and Verifying Large Language Models from Single-Token Output Distributions},
  year = {2026},
  eprint = {2607.10252},
  archivePrefix = {arXiv}
}

@misc{cai2025payfor,
  author = {Cai, W. and Shi, T. and Zhao, X. and Song, D.},
  title = {Are You Getting What You Pay For? Auditing Model Substitution in {LLM} {APIs}},
  year = {2025},
  eprint = {2504.04715},
  archivePrefix = {arXiv}
}

@misc{chen2026sysprompt,
  author = {Chen, L. and Gao, Y. and Wang, J. and Ji, K. and Chen, H.},
  title = {Do System Prompts Leave Behavioral Fingerprints? A Large-Scale Empirical Study of Clone Detection via Output Similarity},
  year = {2026},
  eprint = {2608.24461},
  archivePrefix = {arXiv}
}

@misc{du2026palmyrax6,
  author = {Du, P. and Kamble, K. and Vasudev, R. and Yang, Z. and Nadimpally, R. and Krishna, A. and Alshikh, W. and Bikel, D. M.},
  title = {Palmyra x6 Technical Report: An Agentic, Tool-Use Model Post-Trained via Anchored Supervised Fine-Tuning},
  year = {2026},
  eprint = {2608.16620},
  archivePrefix = {arXiv}
}

@misc{fu2026nondeterminism,
  author = {Fu, T. and Mart{\'\i}nez, G. and Conde, J. and Arriaga, C. and Reviriego, P. and Qi, X. and Liu, S.},
  title = {Beyond Reproducibility: Token Probabilities Expose Large Language Model Nondeterminism},
  year = {2026},
  eprint = {2601.06118},
  archivePrefix = {arXiv}
}

@inproceedings{gao2025met,
  author = {Gao, I. and Liang, P. and Guestrin, C.},
  title = {Model Equality Testing: Which Model Is This {API} Serving?},
  booktitle = {International Conference on Learning Representations (ICLR)},
  year = {2025},
  note = {arXiv:2410.20247}
}

@misc{he2025nondeterminism,
  author = {He, H. and {Thinking Machines Lab}},
  title = {Defeating Nondeterminism in {LLM} Inference},
  year = {2025},
  howpublished = {Thinking Machines Lab blog, 10 September 2025},
  url = {https://thinkingmachines.ai/blog/defeating-nondeterminism-in-llm-inference/}
}

@inproceedings{jiang2025hivemind,
  author = {Jiang, L. and Chai, Y. and Li, M. and Liu, M. and Fok, R. and Dziri, N. and Tsvetkov, Y. and Sap, M. and Albalak, A. and Choi, Y.},
  title = {Artificial Hivemind: The Open-Ended Homogeneity of Language Models (and Beyond)},
  booktitle = {Advances in Neural Information Processing Systems (NeurIPS)},
  year = {2025},
  note = {arXiv:2510.22954}
}

@misc{leshin2026stability,
  author = {Leshin, J. and Shah, M. and Timmis, I. and Kang, D.},
  title = {Behavioral Fingerprints for {LLM} Endpoint Stability and Identity},
  year = {2026},
  eprint = {2603.19022},
  archivePrefix = {arXiv}
}

@misc{parikh2026census,
  author = {Parikh, T.},
  title = {The One-Word Census: Answer-Choice Conformity Across 105 Language Models},
  year = {2026},
  eprint = {2607.12796},
  archivePrefix = {arXiv},
  note = {Version 4}
}

@inproceedings{pasquini2025llmmap,
  author = {Pasquini, D. and Kornaropoulos, E. M. and Ateniese, G.},
  title = {{LLMmap}: Fingerprinting for Large Language Models},
  booktitle = {USENIX Security Symposium},
  year = {2025},
  note = {arXiv:2407.15847}
}

@inproceedings{richardeau2026flips,
  author = {Richardeau, G. and Dashyan, G. and Le Merrer, E. and Tredan, G.},
  title = {{FLIPS}: Instance-Fingerprinting for {LLMs} via Pseudo-random Sequences},
  booktitle = {International Conference on Machine Learning (ICML)},
  year = {2026},
  note = {arXiv:2606.03330}
}

@misc{wang2026agentprov,
  author = {Wang, X. and Zhao, B. and Backes, M. and Boenisch, F. and Dziedzic, A.},
  title = {{AgentProv}: Auditing Agentic {LLM} {API} Providers via Tool-use Policy Probes},
  year = {2026},
  eprint = {2609.00052},
  archivePrefix = {arXiv}
}

@misc{zhang2026iris,
  author = {Zhang, Y. and Zhang, Z.-H. and Qin, H.},
  title = {Which Model Is Actually Serving You? {IRIS}: Budgeted Black-Box Auditing of Model Substitution and Routing Dilution in {LLM} Gateways},
  year = {2026},
  eprint = {2607.20860},
  archivePrefix = {arXiv}
}

@misc{zhu2026unstable,
  author = {Zhu, H. and Zhang, J.},
  title = {Clean Engineering, Unstable Measurement: A Preregistered Reliability Failure of Black-Box {LLM} Observers on Shared Endpoints},
  year = {2026},
  eprint = {2609.04198},
  archivePrefix = {arXiv}
}

@misc{zhu2025rut,
  author = {Zhu, X. and Ye, Y. and Qiu, T. and Zhu, H. and Tan, S. and Mannan, A. and Michala, J. and Popa, R. A. and Neiswanger, W.},
  title = {Auditing Black-Box {LLM} {APIs} with a Rank-Based Uniformity Test},
  year = {2025},
  eprint = {2506.06975},
  archivePrefix = {arXiv}
}

\appendix

\section{The 96 questions}
\label{app:questions}

Every question is ``Name a \{category\}. Reply with one word only.'', sent as the only user turn
with no system prompt, except ``any word'', which is ``Pick a word. Reply with one word only.''
A dagger marks the 31 questions of the original census, used in \S\ref{sec:sysprompt} and
\S\ref{sec:weeks}.

\medskip\noindent{\small
animal$^\dagger$, any word$^\dagger$, author, berry, beverage$^\dagger$, bird$^\dagger$, board game$^\dagger$, brand, bread, building, candy, car brand, cheese$^\dagger$, city$^\dagger$, clothing, cloud, cocktail, color$^\dagger$, company, composer, condiment$^\dagger$, constellation, country$^\dagger$, currency, dance$^\dagger$, deity, desert, dessert$^\dagger$, dinosaur$^\dagger$, dog breed, element, emotion$^\dagger$, fabric$^\dagger$, fish$^\dagger$, flower$^\dagger$, food, fruit$^\dagger$, furniture, gemstone$^\dagger$, genre, grain, hat, herb$^\dagger$, hero, hobby$^\dagger$, holiday, insect$^\dagger$, instrument$^\dagger$, inventor, island, lake, landmark, language$^\dagger$, leader, mammal, material, metal$^\dagger$, mineral, mountain, mushroom, mythical creature$^\dagger$, nut, occupation$^\dagger$, painter, pasta, philosopher, plant, poet, religion, reptile, river, rock, sauce, scientist, sea creature, shape, smell, snack, sound, soup, spice, sport$^\dagger$, superpower, team, texture, tool$^\dagger$, toy, tree$^\dagger$, utensil, vegetable$^\dagger$, vehicle, vice, villain, virtue, weapon, weather.
}

\section{The 94 models}
\label{app:models}

OpenRouter label, family label and collection dates (month-day, 2026). A dagger marks the 14
models of \S\ref{sec:sysprompt} and \S\ref{sec:weeks}; an asterisk marks the 10 models of the blind
test (\S\ref{sec:host}).

{\small
\begin{longtable}{lll}
\toprule
model & family & collected \\
\midrule
\endhead

claude-fable-5 & claude & 07-04 to 09-30 \\
claude-fable-5.1 & claude & 09-03 to 09-30 \\
claude-haiku-4.5 & claude & 07-03 to 09-30 \\
claude-opus-4.1 & claude & 09-25 to 09-30 \\
claude-opus-4.5 & claude & 09-25 to 09-30 \\
claude-opus-4.6 & claude & 09-25 to 09-30 \\
claude-opus-4.7 & claude & 09-25 to 09-30 \\
claude-opus-4.8 & claude & 07-03 to 09-30 \\
claude-opus-5 & claude & 09-03 to 09-30 \\
claude-opus-5.5$^\dagger$ & claude & 09-25 to 09-30 \\
claude-sonnet-4.5 & claude & 09-25 to 09-30 \\
claude-sonnet-4.6 & claude & 07-04 to 09-30 \\
claude-sonnet-5 & claude & 07-03 to 09-30 \\
claude-sonnet-5.5 & claude & 09-28 to 09-30 \\
command-a & cohere & 07-04 to 09-30 \\
deepseek-chat-v3-0324 & deepseek & 07-03 to 09-30 \\
deepseek-v3.2$^\ast$ & deepseek & 07-03 to 09-30 \\
deepseek-v4-flash$^\ast$ & deepseek & 07-03 to 09-30 \\
deepseek-v4-pro$^\dagger$ & deepseek & 09-03 to 09-30 \\
ernie-4.5-vl-424b-a47b & ernie & 07-03 to 09-30 \\
gemini-2.5-flash & gemini & 07-03 to 09-30 \\
gemini-2.5-pro & gemini & 09-25 to 09-30 \\
gemini-3-flash-preview & gemini & 09-25 to 09-30 \\
gemini-3.1-flash-lite & gemini & 09-03 to 09-30 \\
gemini-3.1-pro-preview & gemini & 07-03 to 09-30 \\
gemini-3.5-flash & gemini & 07-03 to 09-30 \\
gemini-3.6-flash & gemini & 09-25 to 09-30 \\
gemini-3.8-flash$^\dagger$ & gemini & 09-03 to 09-30 \\
gemma-3-27b-it$^\ast$ & gemini & 07-03 to 09-30 \\
gemma-4-26b-a4b-it & gemini & 09-03 to 09-30 \\
gemma-4-31b-it$^\dagger$ & gemini & 09-03 to 09-30 \\
glm-4.7 & glm & 07-04 to 09-30 \\
glm-5.3$^\dagger$ & glm & 09-03 to 09-30 \\
glm-5.3-flash & glm & 09-03 to 09-30 \\
gpt-3.5-turbo & gpt & 07-03 to 09-30 \\
gpt-4-turbo & gpt & 07-04 to 09-30 \\
gpt-4.1 & gpt & 07-04 to 09-30 \\
gpt-4o & gpt & 07-04 to 09-30 \\
gpt-4o-mini-2024-07-18$^\dagger$ & gpt & 07-03 to 09-30 \\
gpt-5 & gpt & 07-04 to 09-30 \\
gpt-5.4 & gpt & 07-03 to 09-30 \\
gpt-5.4-mini & gpt & 09-25 to 09-30 \\
gpt-5.5 & gpt & 07-03 to 09-30 \\
gpt-5.6-luna & gpt & 07-10 to 09-30 \\
gpt-5.6-sol & gpt & 07-10 to 09-30 \\
gpt-5.6-terra & gpt & 07-10 to 09-30 \\
gpt-6-astra & gpt & 09-04 to 09-30 \\
gpt-6-luna & gpt & 09-25 to 09-30 \\
gpt-6-sol & gpt & 09-25 to 09-30 \\
gpt-6.1-sol$^\dagger$ & gpt & 09-29 to 09-30 \\
gpt-oss-120b$^\ast$ & gpt-oss & 09-03 to 09-30 \\
gpt-oss-20b & gpt-oss & 09-03 to 09-30 \\
granite-4.2-30b & granite & 10-02 \\
granite-4.2-8b & granite & 10-02 \\
grok-4.20 & grok & 07-10 to 09-30 \\
grok-4.3 & grok & 07-04 to 09-30 \\
grok-4.5 & grok & 07-10 to 09-30 \\
grok-4.6 & grok & 09-03 to 09-30 \\
grok-4.7$^\dagger$ & grok & 09-25 to 09-30 \\
hermes-3-llama-3.1-70b & hermes & 10-02 \\
hermes-4-405b & hermes & 09-29 to 09-30 \\
hy4-preview & hunyuan & 10-02 \\
inkling & inkling & 10-02 \\
inkling-small & inkling & 10-02 \\
kimi-k2 & kimi & 09-25 to 09-30 \\
kimi-k2.5 & kimi & 07-04 to 09-30 \\
kimi-k3$^\dagger$ & kimi & 09-03 to 09-30 \\
ling-3.0-flash & ling & 10-02 \\
llama-3.3-70b-instruct$^\dagger$$^\ast$ & llama & 07-03 to 09-30 \\
llama-4-maverick & llama & 07-03 to 09-30 \\
llama-4-scout$^\ast$ & llama & 09-03 to 09-30 \\
mimo-v2.6-pro & mimo & 10-02 \\
minimax-m3 & minimax & 09-03 to 09-30 \\
mistral-nemo$^\ast$ & mixtral & 09-03 to 09-30 \\
mistral-small-2603$^\dagger$ & mixtral & 09-29 to 09-30 \\
mistral-small-3.2-24b-instruct$^\ast$ & mixtral & 09-03 to 09-30 \\
mixtral-8x22b-instruct & mixtral & 07-03 to 09-30 \\
muse-glimmer-30b & muse & 10-02 \\
muse-spark-1.3$^\dagger$ & muse & 09-25 to 09-30 \\
mythomax-l2-13b & mythomax & 07-04 to 09-30 \\
nemotron-3-super-120b-a12b$^\ast$ & nemotron & 09-29 to 09-30 \\
nemotron-3-ultra-550b-a55b$^\dagger$ & nemotron & 09-29 to 09-30 \\
nemotron-3.5-lightning & nemotron & 09-29 to 09-30 \\
palmyra-x5 & palmyra & 07-04 to 09-30 \\
phi-4 & phi & 10-02 \\
qwen-2.5-72b-instruct & qwen & 07-04 to 09-30 \\
qwen3-235b-a22b-2507$^\ast$ & qwen & 07-03 to 09-30 \\
qwen3.5-27b & qwen & 09-03 to 09-30 \\
qwen3.6-35b-a3b & qwen & 09-03 to 09-30 \\
qwen3.7-plus & qwen & 09-25 to 09-30 \\
qwen3.8-2.4t-a95b$^\dagger$ & qwen & 09-03 to 09-30 \\
qwen3.8-27b & qwen & 09-29 to 09-30 \\
seed-2.0-pro & seed & 10-02 \\
sonar & sonar & 07-04 to 09-30 \\

\bottomrule
\end{longtable}}

The census transcript directories hold 106 models. Twelve are left out. Three have no question
answered eight times (claude-3-haiku, granite-4.1-8b, hermes-4-70b), because they left OpenRouter
before the second four runs. Six answer fewer than 90 questions eight times, from failed cells or
replies that are not one word: wizardlm-2-8x22b (69), qwen3.5-9b (79), hermes-3-llama-3.1-405b
(83), qwen3.5-122b-a10b (84), hunyuan-a13b-instruct (87) and nemotron-3-nano-30b-a3b (88). Three
were added after the panel was fixed on 6 October: claude-haiku-5.5, and deepseek-r1 and
step-3.7-flash, which crossed the threshold after failed cells were refilled on 11 October. The
panel was fixed after the 2 October audit, and the audit was scored against it.

\section{From reply to answer}
\label{app:normalize}

Replies are reduced to answers in six steps (\texttt{analyze.py}).
\begin{enumerate}\setlength{\itemsep}{0pt}
\item Strip a closed reasoning block, chat-template tokens, a leading ``ASSISTANT:'', anything
after a stop marker, and leftover tags. Reasoning returned in the API's separate reasoning field is
never read.
\item Take the first line that is an answer, skipping lead-ins (``Sure, here is one:'') and
asides. A self-correction (``Wait, \ldots{} One word: Stonehenge'') replaces the first answer.
\item Lower-case, fold accents, drop punctuation and emoji, and take the last word, so ``A common
color is blue.'' gives ``blue''. A two- or three-word name ending in a generic head noun of its
category is joined (``soy sauce'' to ``soysauce'', ``golden retriever'' to ``goldenretriever'').
\item Drop failed cells: unclosed reasoning, a reply that opens as leaked reasoning, replies over
15 words, bare acknowledgments, single letters and function words. They are not scored. Of 72,192
panel cells, 83 (0.11\%) give no usable answer; the most are Mixtral 8x22B (23) and Mistral Small
2603 (19).
\item Merge spelling variants with a hand-checked map per category, built from a similarity scan
and a category-by-category review by four reviewers before any fingerprint analysis. Near-matches
that are different things are kept apart (serenity and serendipity, pine and pineapple).
\item Merge plurals: an answer ending in ``s'' becomes its stem when the stem occurs in the panel's
answers to that question.
\end{enumerate}
Answers from outside the panel (the blind test, the audit, the follow-ups) go through the same
steps and are merged onto the panel's answers.

\section{Edge cases in the score}
\label{app:edges}

\begin{itemize}\setlength{\itemsep}{0pt}
\item Profiles use a model's first eight answers per question.
\item A held-out position with no answer (a failed cell) is skipped, so that test set has fewer
answers.
\item A candidate with fewer profiled answers on a question has a smaller $N_m(c)$. Then $V_c$ no
longer cancels exactly across candidates. This is the one place the vocabulary size affects the
ranking. $V_c$ counts answers from all samples, held-out ones included.
\item An answer a candidate never gave scores $\log(0.1 / (N_m(c) + 0.1\,V_c))$. Appendix~\ref{app:scorers}
shows how much the pseudocount matters.
\item Ties go to the alphabetically first label.
\end{itemize}

\section{Scorer comparison}
\label{app:scorers}

The 66 questions every model answered eight times; the folds of \S\ref{sec:method}; 20 draws per
budget; Jensen--Shannon distance in bits, averaged over questions; ties in the most-common-answer
scorer broken at random. Mean exact-model accuracy, standard deviation across draws in
parentheses. The pseudocounts 0.01 and 0.1 give the same accuracy, and 1 costs about five points.
At 40 questions asked once, the likelihood ranges from 89\% to 94\% over the 20 draws.

{\small
\begin{tabular}{lcccccc}
\toprule

scorer & 10 $\times$ 1 & 20 $\times$ 1 & 40 $\times$ 1 & 10 $\times$ 2 & 20 $\times$ 2 & 40 $\times$ 2 \\

\midrule

likelihood 0.01 & 50.3 (7.7) & 76.2 (3.6) & 91.9 (1.2) & 64.7 (8.7) & 88.6 (3.1) & 96.9 (0.5) \\
likelihood 0.1 & 50.1 (7.9) & 77.1 (3.7) & 92.0 (1.3) & 64.9 (9.1) & 89.3 (2.6) & 96.9 (0.5) \\
likelihood 0.5 & 48.0 (8.2) & 74.7 (3.8) & 89.4 (1.7) & 61.4 (8.8) & 86.1 (2.9) & 95.6 (0.9) \\
likelihood 1.0 & 46.1 (8.3) & 72.0 (4.0) & 87.0 (2.1) & 58.0 (8.9) & 83.0 (3.2) & 94.0 (1.2) \\
jsd & 32.8 (7.1) & 51.8 (4.5) & 64.8 (2.7) & 49.9 (8.7) & 71.8 (4.9) & 84.6 (2.7) \\
modal & 31.8 (6.9) & 55.4 (3.6) & 75.8 (1.8) & 39.3 (9.0) & 67.6 (4.2) & 85.3 (2.0) \\

\bottomrule
\end{tabular}}

\section{Question selection}
\label{app:selection}

The objective is the mean log posterior of the true model, a softmax over the 94 candidates,
summed over chosen questions. The first 25 questions chosen in each half, with the accuracy of the
chosen set on the other half.

{\small
\begin{longtable}{rllll}
\toprule
& \multicolumn{2}{l}{chosen on samples 0--3} & \multicolumn{2}{l}{chosen on samples 4--7} \\
$k$ & question & accuracy & question & accuracy \\
\midrule
\endhead

1 & mammal & 15\% & building & 11\% \\
2 & sound & 34\% & reptile & 28\% \\
3 & bird & 47\% & poet & 50\% \\
4 & instrument & 51\% & landmark & 56\% \\
5 & reptile & 61\% & smell & 63\% \\
6 & landmark & 72\% & philosopher & 72\% \\
7 & dessert & 73\% & bird & 78\% \\
8 & clothing & 77\% & painter & 78\% \\
9 & gemstone & 79\% & color & 80\% \\
10 & author & 80\% & dessert & 82\% \\
11 & snack & 85\% & animal & 87\% \\
12 & mythical creature & 85\% & snack & 87\% \\
13 & mushroom & 88\% & mythical creature & 88\% \\
14 & occupation & 90\% & food & 90\% \\
15 & insect & 92\% & villain & 90\% \\
16 & material & 92\% & occupation & 94\% \\
17 & tool & 93\% & element & 95\% \\
18 & dance & 94\% & mushroom & 95\% \\
19 & vegetable & 93\% & material & 95\% \\
20 & inventor & 95\% & dinosaur & 96\% \\
21 & hat & 95\% & instrument & 96\% \\
22 & mountain & 96\% & island & 98\% \\
23 & berry & 96\% & gemstone & 98\% \\
24 & dinosaur & 97\% & clothing & 98\% \\
25 & utensil & 96\% & author & 98\% \\

\bottomrule
\end{longtable}}

\section{The audited endpoints}
\label{app:audit}

Each scored endpoint, by claimed model and a letter. Letters follow the alphabetical order of
the endpoint directories in \texttt{transcripts-providers/}. Price is output dollars per million tokens. Rank is the
claimed model's rank among 94 profiles; margin is its score minus the best other model's, in nats
per question; pinned is the rank against a pinned reference profile (\S\ref{sec:results}); $q$ is
the peer test.

{\scriptsize
\begin{longtable}{llrlrrrr}
\toprule
endpoint & precision & price & identified as & rank & margin & pinned & $q$ \\
\midrule
\endhead

deepseek-v3.2/A & fp8 & 1.11 & deepseek-v3.2 & 1 & +1.63 & 1 & 0.041 \\
deepseek-v3.2/B & fp8 & 0.38 & deepseek-v3.2 & 1 & +1.16 & 1 & 0.292 \\
deepseek-v3.2/C & fp8 & 0.42 & deepseek-v3.2 & 1 & +0.56 & 1 & 1.000 \\
deepseek-v3.2/D & fp4 & 0.38 & deepseek-v3.2 & 1 & +1.27 & 1 & 0.345 \\
deepseek-v3.2/E & unstated & 0.96 & deepseek-v3.2 & 1 & +1.39 & 1 & 0.557 \\
deepseek-v3.2/F & unstated & 1.50 & deepseek-v3.2 & 1 & +1.51 & 1 & 0.006 \\
deepseek-v3.2/G & fp8 & 0.31 & deepseek-v3.2 & 1 & +1.28 & 1 & 0.754 \\
deepseek-v3.2/H & unstated & 1.68 & deepseek-v3.2 & 1 & +1.60 & 1 & 0.979 \\
deepseek-v3.2/I & unstated & 4.50 & deepseek-v3.2 & 1 & +1.54 & 1 & 1.000 \\
deepseek-v3.2/J & unstated & 1.00 & deepseek-v3.2 & 1 & +1.54 & 1 & 0.848 \\
deepseek-v3.2/K & unstated & 4.50 & deepseek-v3.2 & 1 & +1.70 & 1 & 0.848 \\
deepseek-v3.2/L & fp8 & 0.42 & deepseek-v3.2 & 1 & +1.27 & 1 & 1.000 \\
deepseek-v3.2/M & unstated & 0.39 & deepseek-v3.2 & 1 & +1.41 & 1 & 0.024 \\
deepseek-v4-flash/A & fp8 & 0.27 & deepseek-v4-flash & 1 & +0.20 & 1 & 0.848 \\
deepseek-v4-flash/B & fp4 & 0.28 & deepseek-v4-flash & 1 & +0.97 & 1 & 1.000 \\
deepseek-v4-flash/C & unstated & 0.56 & deepseek-v4-flash & 1 & +1.57 & 1 & 0.006 \\
deepseek-v4-flash/D & fp8 & 0.05 & deepseek-v4-flash & 1 & +0.87 & 1 & 1.000 \\
deepseek-v4-flash/E & fp8 & 0.18 & deepseek-v4-flash & 1 & +1.71 & 1 & 0.006 \\
deepseek-v4-flash/F & unstated & 0.20 & deepseek-v4-flash & 1 & +1.35 & 1 & 0.006 \\
deepseek-v4-flash/G & fp8 & 0.18 & deepseek-v4-flash & 1 & +1.43 & 1 & 0.333 \\
deepseek-v4-flash/H & fp8 & 0.50 & deepseek-v4-flash & 1 & +0.57 & 1 & 1.000 \\
deepseek-v4-flash/I & fp8 & 0.28 & deepseek-v4-flash & 1 & +0.12 & 1 & 0.478 \\
deepseek-v4-flash/J & fp8 & 1.60 & deepseek-v4-flash & 1 & +1.33 & 1 & 0.439 \\
deepseek-v4-flash/K & fp8 & 0.28 & deepseek-v4-flash & 1 & +0.24 & 1 & 1.000 \\
deepseek-v4-flash/L & fp4 & 1.28 & qwen3.8-27b & 4 & -0.25 & 7 & 0.048 \\
deepseek-v4-flash/M & fp8 & 0.28 & deepseek-v4-flash & 1 & +1.18 & 1 & 0.006 \\
deepseek-v4-flash/N & fp8 & 0.06 & deepseek-v4-flash & 1 & +0.87 & 1 & 1.000 \\
deepseek-v4-flash/O & unstated & 0.19 & deepseek-v4-flash & 1 & +0.92 & 1 & 1.000 \\
gemma-3-27b-it/A & fp8 & 0.16 & gemma-3-27b-it & 1 & +3.89 & 1 & 0.164 \\
gemma-3-27b-it/B & fp8 & 0.30 & gemma-3-27b-it & 1 & +3.39 & 1 & 0.024 \\
gemma-3-27b-it/C & bf16 & 0.20 & gemma-3-27b-it & 1 & +3.84 & 1 & 0.215 \\
gemma-3-27b-it/D & fp8 & 0.45 & gemma-3-27b-it & 1 & +3.48 & 1 & 0.006 \\
glm-4.7/A & fp8 & 1.85 & glm-4.7 & 1 & +0.85 & 1 & 0.558 \\
glm-4.7/B & fp4 & 1.75 & glm-4.7 & 1 & +0.81 & 1 & 0.173 \\
glm-4.7/C & unstated & 2.20 & glm-4.7 & 1 & +0.63 & 1 & 0.011 \\
glm-4.7/D & fp4 & 2.50 & glm-4.7 & 1 & +0.55 & 1 & 0.128 \\
glm-4.7/E & fp8 & 1.98 & glm-4.7 & 1 & +0.56 & 1 & 0.006 \\
glm-4.7/F & fp4 & 1.93 & glm-4.7 & 1 & +0.47 & 1 & 0.054 \\
glm-4.7/G & fp4 & 2.20 & glm-4.7 & 1 & +0.42 & 2 & 0.006 \\
gpt-oss-120b/A & bf16 & 0.19 & gpt-oss-120b & 1 & +1.65 & 1 & 0.486 \\
gpt-oss-120b/B & unstated & 0.60 & gpt-oss-120b & 1 & +1.70 & 1 & 1.000 \\
gpt-oss-120b/C & unstated & 0.60 & gpt-oss-120b & 1 & +1.96 & 1 & 0.921 \\
gpt-oss-120b/D & fp4 & 0.50 & gpt-oss-120b & 1 & +1.89 & 1 & 0.478 \\
gpt-oss-120b/E & fp16 & 0.75 & gpt-oss-120b & 1 & +1.45 & 1 & 0.139 \\
gpt-oss-120b/F & fp4 & 0.17 & gpt-oss-120b & 1 & +1.73 & 1 & 0.176 \\
gpt-oss-120b/G & bf16 & 0.25 & gpt-oss-120b & 1 & +1.87 & 1 & 0.478 \\
gpt-oss-120b/H & bf16 & 0.17 & gpt-oss-120b & 1 & +1.59 & 1 & 0.322 \\
gpt-oss-120b/I & bf16 & 0.60 & gpt-oss-120b & 1 & +2.00 & 1 & 0.059 \\
gpt-oss-120b/J & bf16 & 0.18 & gpt-oss-120b & 1 & +1.73 & 1 & 1.000 \\
gpt-oss-120b/K & unstated & 0.42 & gpt-oss-120b & 1 & +2.07 & 1 & 0.754 \\
gpt-oss-120b/L & unstated & 0.36 & gpt-oss-120b & 1 & +0.48 & 1 & 0.006 \\
gpt-oss-120b/M & unstated & 0.60 & gpt-oss-120b & 1 & +1.85 & 1 & 1.000 \\
gpt-oss-120b/N & fp8 & 0.28 & gpt-oss-120b & 1 & +1.52 & 1 & 1.000 \\
gpt-oss-120b/O & unstated & 0.75 & gpt-oss-120b & 1 & +1.65 & 1 & 0.205 \\
gpt-oss-120b/P & fp4 & 0.60 & gpt-oss-120b & 1 & +1.78 & 1 & 0.444 \\
gpt-oss-120b/Q & fp4 & 0.25 & gpt-oss-120b & 1 & +1.24 & 1 & 0.716 \\
gpt-oss-120b/R & fp4 & 0.75 & gpt-oss-120b & 1 & +1.95 & 1 & 1.000 \\
gpt-oss-120b/S & unstated & 0.60 & gpt-oss-120b & 1 & +0.96 & 1 & 0.333 \\
gpt-oss-120b/T & unstated & 0.95 & gpt-oss-120b & 1 & +1.85 & 1 & 1.000 \\
gpt-oss-120b/U & fp8 & 0.60 & gpt-oss-120b & 1 & +1.59 & 1 & 0.765 \\
gpt-oss-120b/V & unstated & 0.60 & gpt-oss-120b & 1 & +1.84 & 1 & 0.145 \\
kimi-k2/A & fp8 & 2.30 & kimi-k2 & 1 & +1.05 & -- & -- \\
kimi-k2.5/A & unstated & 3.00 & kimi-k2 & 2 & -0.90 & 2 & 0.006 \\
kimi-k2.5/B & int4 & 2.50 & kimi-k2.5 & 1 & +1.09 & 1 & 1.000 \\
kimi-k2.5/C & unstated & 2.85 & kimi-k2.5 & 1 & +1.11 & 1 & 0.938 \\
kimi-k2.5/D & int4 & 2.25 & kimi-k2.5 & 1 & +1.64 & 1 & 0.119 \\
kimi-k2.5/E & unstated & 3.32 & kimi-k2.5 & 1 & +1.12 & 1 & 0.767 \\
llama-3.3-70b-instruct/A & fp8 & 0.52 & llama-3.3-70b-instruct & 1 & +3.58 & 1 & 0.458 \\
llama-3.3-70b-instruct/B & fp8 & 2.25 & llama-3.3-70b-instruct & 1 & +3.70 & 1 & 1.000 \\
llama-3.3-70b-instruct/C & fp16 & 0.71 & llama-3.3-70b-instruct & 1 & +3.44 & 1 & 0.754 \\
llama-3.3-70b-instruct/D & fp8 & 0.32 & llama-3.3-70b-instruct & 1 & +3.11 & 1 & 0.290 \\
llama-3.3-70b-instruct/E & unstated & 0.79 & llama-3.3-70b-instruct & 1 & +3.01 & 1 & 0.021 \\
llama-3.3-70b-instruct/F & bf16 & 0.40 & llama-3.3-70b-instruct & 1 & +3.57 & 1 & 0.333 \\
llama-3.3-70b-instruct/G & fp8 & 0.50 & llama-3.3-70b-instruct & 1 & +3.52 & 1 & 1.000 \\
llama-3.3-70b-instruct/H & unstated & 0.90 & llama-3.3-70b-instruct & 1 & +3.05 & 1 & 0.145 \\
llama-3.3-70b-instruct/I & unstated & 1.04 & llama-3.3-70b-instruct & 1 & +3.65 & 1 & 0.059 \\
llama-4-maverick/A & unstated & 0.65 & llama-4-maverick & 1 & +4.25 & 1 & 0.006 \\
llama-4-maverick/B & fp8 & 0.85 & gemma-4-31b-it & 91 & -7.01 & 92 & 0.006 \\
llama-4-maverick/C & fp8 & 1.00 & llama-4-maverick & 1 & +4.41 & 1 & 0.006 \\
llama-4-scout/A & fp8 & 0.30 & llama-4-scout & 1 & +4.36 & 1 & 0.333 \\
llama-4-scout/B & bf16 & 0.59 & llama-4-scout & 1 & +4.36 & 1 & 0.179 \\
mistral-small-3.2-24b-instruct/A & fp8 & 0.20 & mistral-small-3.2-24b-instruct & 1 & +1.61 & 1 & 0.700 \\
mistral-small-3.2-24b-instruct/B & unstated & 0.30 & mistral-small-2603 & 5 & -2.37 & 10 & 0.006 \\
mistral-small-3.2-24b-instruct/C & bf16 & 0.30 & mistral-small-3.2-24b-instruct & 1 & +0.80 & 1 & 0.672 \\
mistral-small-3.2-24b-instruct/D & fp8 & 0.25 & mistral-small-3.2-24b-instruct & 1 & +1.61 & 1 & 0.059 \\
nemotron-3-super-120b-a12b/A & bf16 & 0.40 & nemotron-3-super-120b-a12b & 1 & +0.89 & 1 & 0.399 \\
nemotron-3-super-120b-a12b/B & fp8 & 0.45 & nemotron-3-super-120b-a12b & 1 & +1.33 & 1 & 0.345 \\
qwen3-235b-a22b-2507/A & unstated & 0.60 & qwen3-235b-a22b-2507 & 1 & +1.98 & 1 & 0.006 \\
qwen3-235b-a22b-2507/B & fp8 & 0.55 & qwen3-235b-a22b-2507 & 1 & +2.06 & 1 & 0.011 \\
qwen3-235b-a22b-2507/C & fp8 & 0.35 & qwen3-235b-a22b-2507 & 1 & +2.03 & 1 & 0.754 \\
qwen3-235b-a22b-2507/D & unstated & 1.00 & qwen3-235b-a22b-2507 & 1 & +1.79 & 1 & 0.091 \\
qwen3-235b-a22b-2507/E & fp8 & 0.60 & qwen3-235b-a22b-2507 & 1 & +1.94 & 1 & 0.439 \\
qwen3-235b-a22b-2507/F & fp8 & 0.58 & qwen3-235b-a22b-2507 & 1 & +1.83 & 1 & 0.478 \\
qwen3-235b-a22b-2507/G & fp8 & 0.80 & qwen3-235b-a22b-2507 & 1 & +1.93 & 1 & 0.439 \\
qwen3-235b-a22b-2507/H & unstated & 0.84 & qwen3-235b-a22b-2507 & 1 & +1.91 & 1 & 0.489 \\
qwen3-235b-a22b-2507/I & fp8 & 0.75 & qwen3-235b-a22b-2507 & 1 & +2.05 & 1 & 0.979 \\

\bottomrule
\end{longtable}}

\end{document}